\documentclass[conference]{IEEEtran}

\IEEEoverridecommandlockouts
\usepackage{amsmath,amssymb,amsfonts}
\usepackage{algorithm,algorithmic}
\usepackage{algorithmic}
\usepackage{graphicx}
\usepackage{hyperref}
\usepackage{amsmath}
\usepackage{textcomp}
\usepackage{float}
\usepackage{booktabs}
\usepackage{bm}
\usepackage{authblk}
\usepackage{threeparttable} 
\usepackage[maxnames=5,maxcitenames=1
]{biblatex}
\usepackage{xcolor}
\def\BibTeX{{\rm B\kern-.05em{\sc i\kern-.025em b}\kern-.08em
    T\kern-.1667em\lower.7ex\hbox{E}\kern-.125emX}}
    
\usepackage{etoolbox}
\newtoggle{final}

\newcommand{\pw}[1]{\iftoggle{final}{#1}{{\color{black} #1}}}
\newcommand{\oywbyisenw}[1]{\iftoggle{final}{#1}{{\color{black} #1}}}

\newcommand{\MAGIC}{MAGIC}
\newcommand{\baseline}{policy rollout}
\newcommand{\Baseline}{Policy rollout}

\begin{document}

\title{Improving Generalization  of Deep Reinforcement Learning-based TSP Solvers\\
\thanks{$*$ Equal contribution. $\dag$  Corresponding author.}
}

\author[\space1\space2]{Wenbin Ouyang$^*$}
\author[\space1\space2]{Yisen Wang$^*$}
\author[1]{Shaochen Han}
\author[1]{Zhejian Jin}
\author[\space1\space3]{Paul Weng (IEEE Senior Member)$\dag$}
\affil[1]{\textit{UM-SJTU Joint Institute, Shanghai Jiao Tong University}}
\affil[2]{\textit{EECS Department, University of Michigan}}
\affil[3]{\textit{Department of Automation, Shanghai Jiao Tong University}\authorcr oywenbin@umich.edu, yisenw@umich.edu.cn, sc.han@sjtu.edu.cn,\authorcr jinzhejian@sjtu.edu.cn, paul.weng@sjtu.edu.cn}

\renewcommand\Authands{ and }

\maketitle

\begin{abstract}
Recent work applying deep reinforcement learning (DRL) to solve traveling salesman problems (TSP) has shown that DRL-based solvers can be fast and competitive with TSP heuristics for small instances, but do not generalize well to larger instances.
In this work, we propose a novel approach named \MAGIC{} that includes a deep learning architecture and a DRL training method.
Our architecture, which integrates a multilayer perceptron, a graph neural network, and an attention model, defines a stochastic policy that sequentially generates a TSP solution.
Our training method includes several innovations: (1) we interleave DRL policy gradient updates with local search \pw{(using a new local search technique)}, (2) we use a  novel simple baseline, and (3) we \pw{apply} curriculum learning.
Finally, we empirically demonstrate that \MAGIC{} is superior to other DRL-based methods on random TSP instances, both in terms of performance and generalizability.
Moreover, our method compares favorably against TSP heuristics and other state-of-the-art approach in terms of performance and computational time.
\end{abstract}


\section{Introduction}\label{sec:introduction}



Traveling Salesman Problem (TSP) is \pw{one of the most} famous combinatorial optimization \pw{problems}. 
Given the coordinates of some points, the \pw{goal in} the TSP problem is to find a shortest \pw{tour} \pw{that} visits each point exactly once and returns to the starting point. 
TSP is an NP-\pw{hard} problem \cite{PAPADIMITRIOU1977237}\pw{, even in its symmetric 2D Euclidean version, which is this paper's focus}. 
Traditional approaches to solve TSP can be classified as exact \pw{or heuristic.} 
Exact solvers, such as Concorde \cite{CC} 
\pw{or based on integer linear programming, can} find \pw{an optimal solution}. 
However, since TSP is NP-\pw{hard}, such algorithms \pw{have} computational times \pw{that} increase exponentially with the size of \pw{a} TSP \pw{instance}. 
\pw{In contrast,} heuristic approaches \pw{provide a} TSP solution with a much shorter computational time \pw{compared to} exact solvers\pw{, but do not guarantee optimality}. 
\pw{These approaches are either constructive (e.g., \pw{farthest} insertion  \cite{attention}), perturbative (e.g., 2-opt \cite{2opt}, LKH \cite{LKH}), or hybrid.}
\pw{However, they may not provide any good performance guarantee and are still computationally costly.
Indeed, even a quadratic computational complexity may become prohibitive when dealing with large TSP instances (e.g., 1000 cities).}

\pw{Thus, r}ecent research \pw{work has} focused on using Deep Learning (DL) to \pw{design faster heuristics to} solve TSP problems.
\pw{Since training on large TSP instances is costly, generalization is a key factor in such DL-based approaches.}
\pw{They are either based on} Supervised Learning (SL) \cite{45283,joshi2019efficient,Fu} \pw{or} Reinforcement Learning (RL) \cite{DBLP:journals/corr/BelloPLNB16,attention,GPN,DBLP:journals/corr/abs-2004-01608}. 
These different approaches\pw{, which are either constructive, perturbative, or hybrid,} have different pros and cons. 
For example, \pw{\textcite{Fu}}'s 
model\pw{, which combines DL with Monte Carlo Tree Search (MCTS) \cite{MCTS}, has} great generalization capabilities. 
Namely, they can train on small TSP instances 
and perform well on larger instances. 
However, \pw{the computational cost of \textcite{Fu}'s model is high due to MCTS}. 
\pw{In contrast, o}ther models 
\pw{(e.g., \cite{joshi2019efficient,attention})}
can solve small TSP instances with fast speed and great performance, but they lack generalizability. 


\pw{In this paper}, we propose \pw{a novel deep RL approach that can achieve excellent performance with good generalizability for a reasonable computational cost.}
%
%
\pw{The contributions of this paper can be summarized as follows.
Our approach is based on an encoder-decoder model (using Graph \pw{Neural} Network (GNN) \cite{GNN} \pw{and} Multilayer Perceptron (MLP) \cite{MLP} as the encoder and an attention mechanism \cite{attention_mechanism} as the decoder), which is trained with a new deep RL method that interleaves policy gradient updates (with a simple baseline called \baseline{} baseline) and local search (with a novel combined local search technique).
Moreover, curriculum learning is applied to help with training and generalization.
Due to all the \pw{used} techniques, we name our model as \MAGIC{} (MLP for {\bf M}, Attention for {\bf A}, GNN for {\bf G}, \pw{Interleaved} local search for {\bf I}, and Curriculum Learning for {\bf C}).}
\pw{Finally, we empirically} \pw{show} that \MAGIC{} is \pw{a} state-of-the-art deep \pw{RL} solver for TSP, which \pw{offers a good trade-off in terms of performance, generalizability, and computational time}. 

This \pw{paper} is structured as follows. 
Section~\ref{sec:related} \pw{overviews related work}. 
\pw{S}ection~\ref{sec:background} \pw{recalls} the \pw{necessary} background. 
Section~\ref{sec:model} introduces our model architecture. \pw{S}ection~\ref{sec:algorithm} describes \pw{our novel training technique by explaining} how we apply local search, \pw{the} \baseline{} baseline, and curriculum learning \pw{during training}. 
\pw{Section~\ref{sec:expe} presents the experimental results and Section~\ref{sec:conclusion} concludes.}

\section{Related Work} \label{sec:related}


\pw{
RL can be used as a constructive heuristic to generate a tour or as a machine learning method integrated in a traditional method, such as \cite{DBLP:journals/corr/abs-2004-01608}, which learns to apply 2-opt.
For space reasons, we mainly discuss deep RL work in the constructive approach (see \cite{Bai} for a more comprehensive survey), since they are the most related to our work. Besides, a recent work \cite{joshi_learning_2020} suggests that RL training may lead to better generalization than supervised learning.

Such deep RL work started with}
Pointer Network \cite{45283}\pw{, which} was proposed \pw{as} a general model \pw{that} could solve an entire class of \pw{TSP} instances. 
\pw{It has an encoder-decoder architecture, both based on recurrent neural networks, combined with an} attention \pw{mechanism} \cite{Bahdanau2015}.
\pw{The model is trained in a supervised way using} solutions generated
by Concorde \cite{CC}. 
The results \pw{are promising}, but \pw{the authors} focused only on small-scale TSP \pw{instances} (\pw{with up to 50 cities}) and did not \pw{deal with} generalization.

\pw{This approach was extended to the RL setting \cite{DBLP:journals/corr/BelloPLNB16} and shown to scale to TSP with up to 100 cities. 
The RL training is based on an actor-critic scheme using tour lengths as unbiased estimates of the value of a policy.
}
\pw{
In contrast to \cite{DBLP:journals/corr/BelloPLNB16}, a value-based deep RL \cite{dai_learning_2017} was also investigated to solve graph combinatorial optimization problems in general and TSP in particular.
The approach uses graph embeddings to represent partial solutions and RL to learn a greedy policy.
}

\pw{
The Attention Model \cite{attention} improves the Pointer Network \cite{45283} notably by replacing the recurrent neural networks by attention models \cite{attention_mechanism} and using \pw{RL training} with a simple greedy rollout baseline. 
These changes allowed them to achieve better results on small-scale TSP instances, as well as to generalize to \pw{100-city TSP instances}. 
However, their model fails to generalize \pw{well} to large-scale TSP (e.g., with 1000 \pw{cities}) and their algorithm does not scale well in terms of memory usage.

A similar, although slightly more complex, approach is proposed in \cite{deudon_learning_2018}, which also suggests to improve the tour returned by the deep RL policy with a 2-opt local search, which makes the overall combination a hybrid heuristics.
In contrast to that work, we not only apply local search as a final improvement step, but also integrate local search in the training of our deep RL model.
Moreover, we use a more sophisticated local search.}

\pw{Moreover, the \pw{Graph Pointer Network (GPN)} model \cite{GPN} was proposed to improve over previous models by exploiting graph neural networks \cite{GNN} and using a central self-critic baseline, which is a centered greedy rollout baseline.
Like \cite{deudon_learning_2018}, 2-opt is also considered.}
As a result, they \pw{report good} results when generalizing to large-scale TSP \pw{instances}. 
\pw{Our simpler model and new training method outperforms GPN on both small and larger TSP instances.
}

\section{Background}\label{sec:background}

\pw{This section provides the necessary information to understand our model architecture (Section~\ref{sec:model}) and our training method} (Section~\ref{sec:algorithm}).
\pw{For any $n\in\mathbb N$, $[n]$ denotes $\{1, 2, \ldots, n\}$. Vectors and matrices are denoted in bold.}

\subsection{Traveling Salesman Problem}

\pw{A \textit{Traveling Salesperson Problem} (TSP) can informally be stated as follows.
Given $N$ cities, the goal in a TSP instance is to find a shortest tour that visits each city exactly once.
Formally, the set of $N$ cities can be identified to the set $[N] = \{1, 2, \ldots, N\}$.
In the symmetric 2D Euclidean version of the TSP problem, each city $i\in [N]$ is characterized by its 2D-coordinates $\bm x_i \in \mathbb R^2$.
Let $X$ denote the set of city coordinates $\{\bm x_i \mid i \in [N]\}$ and $\bm X \in \mathbb R^{N\times 2}$ the matrix containing all these coordinates.
The distance $d_{ij}$ between two cities $(i, j) \in [N]^2$ is usually measured in terms of the L2-norm $\|\cdot\|_2$:
\begin{equation}
    d_{ij}=d_{ji}=\|\bm x_i - \bm x_j\|_2.
\label{eq:distance}
\end{equation}
A feasible TSP solution, called a \textit{tour}, corresponds to a permutation $\sigma$ over $[N]$.
Its length is defined as:
\begin{equation}
    \begin{split}
L_{\sigma}(X) & = \sum_{t=1}^{N} d_{\sigma(t)\sigma(t+1)}  = \sum_{t=1}^{N} \|\bm x_{\sigma(t)} - \bm x_{\sigma(t+1)}||_2
\end{split}
\end{equation}
where for $t \in [N]$, $\sigma(t) \in [N]$ is the $t$-th city visited in the tour defined by $\sigma$, and by abuse of notation, $\sigma(N+1) = \sigma(1)$.
Therefore, the TSP problem can be viewed as the following optimization problem:
\begin{equation}
    \begin{split}
\min_\sigma L_{\sigma}(X) = \min_\sigma \sum_{t=1}^{N} \|\bm x_{\sigma(t)} - \bm x_{\sigma(t+1)}\|_2.
\end{split}
\end{equation}
Since scaling the city positions does not change the TSP solution, we assume in the remaining of the paper that the coordinates of all cities are in the square $[0, 1]^2$, as done in previous work \cite{DBLP:journals/corr/BelloPLNB16,DBLP:journals/corr/abs-2004-01608,attention,GPN}. 
}

\begin{figure}[tb]
\vspace{-0.1cm}
\centering
\includegraphics[scale=0.35]{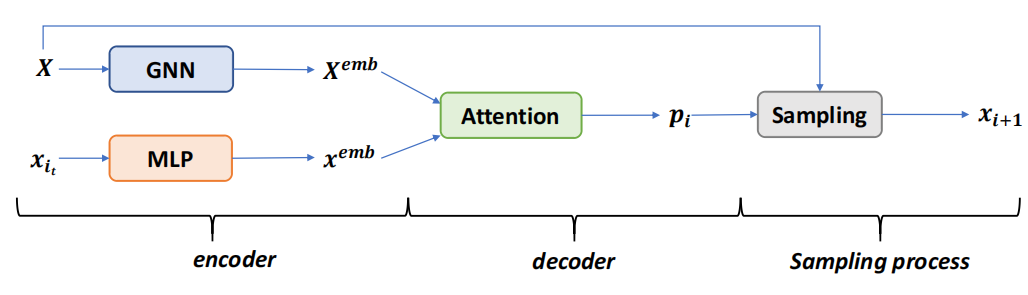}
\vspace{-0.4cm}
\caption{\oywbyisenw{Architecture of the model used in \MAGIC{}.} }
\label{fig:MA}
\vspace{-0.3cm}
\end{figure}

\subsection{Insertion heuristic\pw{s} and k-opt optimization for TSP} \label{sec:heuristics}

\pw{
Since TSP is an NP-hard problem \cite{PAPADIMITRIOU1977237}, various heuristic techniques have been proposed to quickly compute a solution, which may however be sub-optimal.
We recall two family of heuristics: insertion heuristics \cite{attention} and k-opt \cite{kopt}.

Insertion heuristics (including nearest, farthest\pw{,} and random insertion) \pw{are constructive, i.e., they iteratively build a solution. 
They} work as follows.}
They first randomly choose \pw{a} starting \pw{city} \pw{and repeatedly} insert one \pw{new city} at a time \pw{until obtaining} a complete \pw{tour}. 
\pw{Let $\hat\sigma$ denote a partial tour, i.e., a partial list of all cities.} 
Different insertion \pw{heuristics follow} different rule\pw{s} to choose a new \pw{city} $\ell$: 
random insertion choose\pw{s} a new \pw{city} $\ell$ randomly; 
nearest insertion \pw{chooses according to}:
\begin{equation}
    \ell = \arg\min_{j\notin \hat\sigma} \min_{i\in\hat\sigma} d_{ij},
\end{equation}
and \pw{farthest} insertion \pw{chooses according to the following rule}:
\begin{equation}
    \ell = \arg\max_{j\notin \hat\sigma} \min_{i\in\hat\sigma} d_{ij}
\end{equation}
\pw{where $j\not\in\hat\sigma$ means city $j$ is not in the partial tour $\hat\sigma$ and $i \in\hat\sigma$ means city $i$ is in the partial tour.} 
\pw{The position $t$ where city} $\ell$ is inserted into $\pw{\hat\sigma}$ is determined such that: $d_{\hat\sigma(t)\ell} + d_{\ell\hat\sigma(t+1)} - d_{\oywbyisenw{\hat\sigma(t)}\hat\sigma(t+1)}$ is minimized. 

\pw{A classic local search} heuristic \pw{is $k$-opt, which} aims to improve an exist\pw{ing tour} $\sigma$  \pw{by} swapping $k$ chosen edge\pw{s} at each iteration. 
The simplest one is $2$-opt, which \pw{can} replace $\sigma = \big(\sigma(1), \sigma(2)...,\sigma(\pw{i}),...,\sigma(\pw{j}),...,\sigma(N)\big)$
by $\sigma' = \big(\sigma(1),...,\sigma(\pw{i}),\sigma(\pw{j}),\sigma(\pw{j}-1),...,\sigma(\pw{i}+1),\sigma(\pw{j}+1),...,\sigma(N)\big)$ where $\pw{i}<\pw{j}<N$ \pw{if $L_{\sigma'}(X) < L_{\sigma}(X)$.}
\pw{This kind improvement can be found in different ways.
For instance, traditional 2-opt may examine all pairs of edges, while random 2-opt examines randomly-selected pairs.}
\oywbyisenw{
LKH \cite{LKH} is one algorithm that applies $k$-opt and achieve\pw{s} nearly optimal results. 
However, LKH has \pw{a} long run time, especially for large-scale TSP problems.}

\section{Model \pw{and} Architecture}\label{sec:model}

\pw{RL can be used as a constructive method to iteratively generate a complete tour:
at each iteration $t$, a new city $a_t$ with coordinates $\bm x_{a_t}$ is selected based on the list of previously selected cities and the description of the TSP instance.
Formally, this RL model is defined as follows.
A state $\bm s_t$ is composed of the TSP description and the sequence of already visited cities $(i_1, i_2, \ldots, i_{t-1})$ at time step $t$.
State $s_1$ denotes the initial state where no city has been selected yet and state $s_{N+1}$ represents the state where the whole tour has already been constructed.
An action $a_t \in [N]$ corresponds to the next city $i_{t}$ to be visited, i.e., $a_t = i_{t}$.
This RL problem corresponds to a repeated $N$-horizon sequential decision-making problem where the action set for any time step depends on the current state and only contains the cities that have not been visited yet.
The immediate reward for performing an action in a state is given as the negative length between the last visited city and the next chosen one:
\begin{equation}
r(\bm{s}_{t}, a_t) = \left\{
\begin{array}{ll}
0     & \mbox{for } t=1\\
-d_{i_{t-1} i_{t}} & \mbox{for } t = 2, \ldots, N 
\end{array}\right. \label{eq:reward}
\end{equation}
After choosing the first city, no reward can be computed yet.
After the last city, a final additional reward is provided given by $r(\bm s_{N+1}) = - d_{i_N i_1}$.
Thus, a complete trajectory corresponds to a tour and the return of a trajectory is equal to the negative the length of that tour.
Most RL-based constructive solver is based on this RL formulation.
In Section~\ref{sec:algorithm}, we change the return provided to the RL agent to improve its performance using local search.

To perform the selection of the next city, we propose t}he \MAGIC{} \pw{architecture (see Fig.~\ref{fig:MA}), which corresponds to a stochastic policy (see Section~\ref{sec:algorithm} for more details).
It is composed of three parts: (A) an encoder implemented with a graph neural network (GNN) \cite{GNN} and a multilayer perceptron (MLP), (B) a decoder based on an attention mechanism \cite{attention_mechanism}, and (C) a sampling process.}

\subsection{Encoder} \label{sec:encoder}
When solving a TSP problem, not only \pw{should} the last selected city be considered, but also the whole city list should be taken into account as background information. 
\pw{Since the information contained in 2D coordinates is limited and does not include the topology of the cities, we leverage GNN and MLP to encode city coordinates into a higher dimensional space, \pw{depicted} in Fig.~\ref{fig:MA}.}
\oywbyisenw{The GNN is used to encode the city coordinates $\bm X \in \mathbb{R}^{N\times2}$ \pw{into} $\bm X^{emb} \in \mathbb{R}^{N\times H}$ \pw{where $H$ is the dimension of the embedding space}. 
The MLP is used to encode the last selected city $\bm x_{i_t} \in X$ at iteration $t \in [N]$ \pw{into} $\bm x^{emb} \in \mathbb{R}^{H}$.} 
Therefore, generally speaking, \pw{the} GNN and MLP in \MAGIC{} can be viewed as two functions:
\begin{equation}
   \text{GNN}:\mathbb{R}^{N\times2}\rightarrow\mathbb{R}^{N\times H},\ \text{MLP}:\mathbb{R}^2\rightarrow\mathbb{R}^H.
  \label{eq:encoders}
\end{equation}


\begin{figure}[t]
\vspace{-0.1cm}
\centering
\includegraphics[scale=0.4]{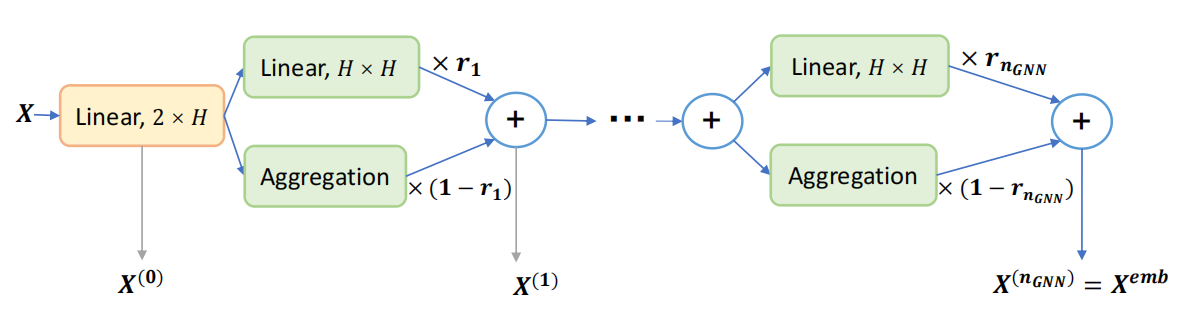}
\vspace{-0.3cm}
\caption{Detailed Architecture of GNN.}
\label{fig:GNN}
\vspace{-0.4cm}
\end{figure}
\subsubsection{GNN} 
GNN is a \pw{technique} which can embed all nodes in a graph together. 
\pw{Similarly} to the GPN model \cite{GPN}, we \pw{use a} GNN to encode the whole city list \pw{of a TSP instance}. 
Fig.~\ref{fig:GNN} shows the detailed architecture of the GNN used in  \MAGIC{}. 
After $\bm X \in \mathbb{R}^{N\times2}$ is transformed \pw{in}to a vector $\bm X^{(0)}\in\mathbb{R}^{N\times H}$, $\bm X^{(0)}$ will go through \pw{$n_{\text{GNN}}$} layers of GNN. 
Each layer of GNN can be expressed as
\begin{equation}
    \bm X^{(t)} = r\cdot\bm X^{(t-1)}\bm\Theta^{(t)} +(1-r)\cdot\mathrm{F}^{(t)}\left(\frac{\bm X^{(t-1)}}{N-1}\right)
    \label{GNN_layer}
\end{equation}
\pw{where $\bm X^{(t-1)} \in \mathbb{R}^{N\times H}$ is the input of the $t^{th}$ layer of the GNN for $t \in [n_{\text{GNN}}]$, $\bm X^{(n_{\text{GNN}})} = \bm X^{emb}$, }
$\bm{\Theta}^{(t)}$ is an $H\times H$ learnable matrix, 
which is represented by a neural network, 
$\mathrm{F}^{(t)}:\mathbb{R}^{N\times H} \rightarrow \mathbb{R}^{N\times H}$ is the aggregation function \cite{GNN} , 
and $\pw{r \in [0, 1]}$ is a trainable parameter. 


\subsubsection{MLP} \pw{While the} GNN provides us with general information within the whole city list $X$\pw{, we also} need to 
\pw{encode} the last selected city $\bm x_{i_t} \in \mathbb{R}^2$. 
\pw{In contrast to previous work using complex architectures like GNN or LSTM \cite{LSTM}, we simply use an MLP. 
Using a GNN would make the embedding of the last selected city depend on the whole city list included the already-visited cities, while using an LSTM would make the embedding depends on the order of visited cities, which is in fact irrelevant.}

\subsection{Decoder}

The decoder of \pw{the \MAGIC{}} model is based on an attention mechanism\pw{, which was also}
\oywbyisenw{used in several previous studies \cite{DBLP:journals/corr/BelloPLNB16,attention,GPN,Fu}.}
The output of \pw{the decoder} is a pointer vector $\bm u^{(t)}\in\mathbb{R}^N$ \cite{DBLP:journals/corr/BelloPLNB16}, which can be expressed as:
\begin{equation}
\label{att_mac}
    \bm{u}^{(t)}_{j} = \begin{cases}-\infty  \text {\ \ \ \ \ \ \ \ \ if}\,\,\exists\, k \le t\text{,}\,\, i_k=j& \\
    \bm w\cdot \tanh \left( \bm X^{emb}_{j}\bm \Theta_{g} + \bm \Theta_{m} \bm x^{emb}\right) \text{ otherwise}&  \end{cases}
\end{equation}
where ${\bm u}^{(t)}_{j}$ is the $j^{th}$ entry of the vector ${\bm u}^{(t)}$, 
$\bm X^{emb}_{j}$ is the $j^{th}$ row of the matrix $\bm X^{emb}$, 
$\bm \Theta_{g}$ and $\bm \Theta_{m}$ are trainable matrices with shape $H\times H$, 
$\bm w \in \mathbb{R}^N$ is a trainable weight vector. 
For the definitions of $\bm X^{emb}$ and $\bm x^{emb}$, please refer to Fig.~\ref{fig:MA}.

\pw{A softmax transformation is used to turn $\bm u^{(t)}$ into a probability distribution $\bm p^{(t)} = \big({\bm p}^{(t)}_{j}\big)_{j \in [N]}$ over cities:}
\begin{equation}
    \bm p^{(t)} = \mathrm{softmax}(\bm u^{(t)}) = \left( \frac{e^{\bm u^{(t)}_{j}}}{\sum_{j=1}^{N} e^{\bm u^{(t)}_{j}}} \right)_{j \in [N]}
    \label{sm}
\end{equation}
where ${\bm p}^{(t)}_{j}$ is the $j^{th}$ entry of the probability distribution ${\bm p}^{(t)}$ and ${\bm u}^{(t)}_{j}$ is the $j^{th}$ entry of the vector ${\bm u}^{(t)}$. 
Notice that if the $j^{th}$ city is visited, then ${\bm u}^{(t)}_{j}=-\infty$ due to \eqref{att_mac}. 
Under this circumstance, ${\bm p}^{(t)}_{j}=0$ according to  \eqref{sm}. 
That is to say, all visited cities cannot be visited again.

\subsection{Sampling Process}

After we obtain the probability distribution ${\bm p}^{(t)}$, it is trivial to select the next city. 
\pw{Indeed,} ${\bm p}^{(t)}$ \pw{corresponds to} the \pw{RL} policy \pw{$\pi_{\bm\theta}$ at time step $t$}:
\begin{equation}
\pi_{\bm\theta}(a_t \mid \bm{s}_t)=\bm p^{(t)}_{a_t} \label{eq:policy}
\end{equation}
where 
$\bm s_t$ (resp. $a_t$) is the \pw{state (resp. action) at time step $t$} and
$\bm p^{(t)}_{a_t}$ is the probability of choosing $a_t$ as the $t^{th}$ city.
Therefore, we just need to sample the next city 
according to the probability distribution 
${\bm p}^{(t)}$.

\section{Algorithm and Training}\label{sec:algorithm}

\pw{
For the training of \MAGIC{}, we propose to interleave standard policy gradient updates with local search.
In contrast to previous work, our idea is to learn a policy that can generate tours that can be easily improved with local search.
Next, we explain our local search technique, which include a novel local insertion-based heuristics.
Then, we present how policy gradient with a simple policy rollout baseline can be applied.
Finally, we motivate the use of stochastic curriculum learning method in our setting.
}

\begin{algorithm}[t]
  \caption{\pw{Local} Insertion Optimization Algorithm}
  \begin{algorithmic}[1]
  \STATE {\bf Input}: A \pw{set} of city coordinates $X=\{\bm x_i\}_{i=1}^N$, current \pw{tour} $\sigma$, and $\gamma \in [0,1]$
  \STATE {\bf Output}: An improved \pw{tour} $\sigma$
  \FOR{$t=1$ \TO $N$}
    \STATE $t^* = \arg\min_{t', \,\,|t'-t|<\gamma\times N} L_{\sigma_{t,t'}}(X)$
    \STATE $\sigma\leftarrow \sigma_{t,t^*}$
  \ENDFOR
  \RETURN $\sigma$
  \end{algorithmic}
\label{alg:IO}
\end{algorithm}

\subsection{Local search}

\pw{We describe the local search technique that we use for training our model and to improve the tour output by the RL policy.
Our technique uses two local search heuristics in combination: random opt and a local insertion heuristics, which is novel to the best of our knowledge.
The two heuristics have been chosen and designed to be computationally efficient, which is important since we will apply them during RL training.}
%
\pw{The motivation for combining two heuristics} is 
that when one method \pw{gets stuck in} some local minimum, the other method \pw{may} help \pw{escape from it}. 

For random 2-opt, we randomly pick 2 arcs for improvement and repeat for $\alpha \times N^{\beta}$, where $N$ is the number of the cities and $\alpha > 0$ and $\beta \in \mathbb R$ are two hyperparameters. 
We set $\alpha$ and $\beta$ here to have a flexible control of the strength of this local search and make it stronger 
\pw{if needed for} larger TSP problems.
\pw{With this procedure, random 2-opt can be much faster than traditional 2-opt.}

Inspired by the insertion heuristics, we propose \textit{\pw{local} insertion optimization}. 
Let \pw{$\sigma$ be the current \pw{tour}} and $\sigma_{t,t'} = (\sigma(1), ..., \sigma(t'), \sigma(t), \sigma(t'+1), ...,\sigma(t-1), \sigma(t+1),...,\sigma(N))$ if $t'\neq t-1$ and $\sigma_{t,t-1} = \sigma$. 
This method (see Algorithm~\ref{alg:IO}) first iterates through all \pw{indices $t \in [N]$}, and for each index $t$, we let $t^* = \arg\min_{t', \,\,|t'-t|<\gamma\times N}L_{\sigma_{t,t'}}(X)$, where $\gamma \in [0,1]$ is a hyperparameter, and then \pw{replace} $\sigma$ by $\sigma_{t,t^*}$. 
\pw{The rationale for restricting the optimization with hyperparameter $\gamma$ is as follows.
For a good suboptimal tour $\sigma$,  cities that are close in terms of visit order in $\sigma$ are usually also close in terms of distance.
In that case, $\sigma_{t,t'}$ is unlikely to improve over $\sigma$ when $t$ and $t'$ are far apart.
Thus}, we set $\gamma$ to limit the search range to increase the \pw{computational} efficiency \pw{of this heuristics}. 

\pw{We call our local search technique \textit{combined local search} (see Algorithm~\ref{alg:CLO}), which}
\pw{applies} random 2-opt followed by \pw{local} insertion optimization repeatedly for $I$ times, where $I$ is a hyperparameter. 
\begin{algorithm}[t]
  \caption{Combined Local Search Algorithm}
  \begin{algorithmic}[1]
  \STATE {\bf Input}: A \pw{set} of city \pw{coordinates} $X=\{x_i\}_{i=1}^N$, current \pw{tour} $\sigma$, hyperparameters $\alpha$, $\beta$, $\gamma$ and $I$ for local search.
  \STATE {\bf Output}: An improved \pw{tour} $\sigma$
  \FOR{$t=1$ \TO $I$}
\pw{
    \FOR{$t'=1$ \TO $\alpha N ^\beta$}
    \STATE $\sigma\leftarrow$ apply random 2-opt on $\sigma$
    \ENDFOR
    \STATE $\sigma\leftarrow$ apply Local Insertion Optimization$(X, \sigma, \gamma)$}
  \ENDFOR
  \RETURN $\sigma$
  \end{algorithmic}
\label{alg:CLO}
\end{algorithm}

\subsection{\pw{Interleaved RL training} with \pw{the} \baseline{} baseline} \label{sec:RL}

\pw{Our model is trained with the REINFORCE \cite{Williams92simplestatistical} algorithm.
The novelty is that we interleave local search with the policy gradient updates.
When the current policy $\pi_{\bm\theta}$ outputs a tour $\sigma$, this solution is further improved with our combined local search technique to obtain a new tour $\sigma_+$.
In contrast to previous work, this tour $\sigma_+$ instead of $\sigma$ is used to evaluate policy $\pi_{\bm\theta}$.
The rationale for this procedure is to make the RL policy and local search work in synergy by favoring learning policies that generate tours that can be easily improved by the combined local search.
If the RL training and local search are not coordinated, as done in previous work, then a trained policy may generate tours that are hard to improve by local search.}



\subsubsection{Policy Gradient} \label{sec:policy_g}

\pw{We recall first standard policy gradient and then explain how we modify it.}
\pw{With} the reward function in \eqref{eq:reward}, 
\pw{the RL goal would be to find} ${\bm\theta}^{\star}$ such that 
\begin{equation}
{\bm\theta}^{\star} = \arg\max _{\bm\theta} J(\bm\theta) = \arg\max _{\bm\theta} \mathbb E_{\oywbyisenw{\tau \sim \bm p_{\bm\theta}(\tau)}}\left[ \sum_{t=1}^{N+1}
r_t\right], 
\label{1515}
\end{equation}
where $r_t=r(\bm{s}_{t}, a_{t})$ for $t\in[N]$, $r_{N+1} = r(\bm{s}_{N+1})$, \pw{$\tau = (\bm s_1, a_1, \bm s_2, a_2, \ldots, \bm s_N, a_N, \bm s_{N+1})$ is a trajectory, and $\bm p_{\bm\theta}$ is the probability distribution over tours induced by policy $\pi_{\bm\theta}$.}
\pw{Recall the} gradient of $J(\theta)$
\cite{Williams92simplestatistical}
\pw{is:}
\begin{align}
\nabla_{\bm\theta} J(\bm\theta) =\mathbb E_\tau \left[\left(\sum_{t=1}^{N} \nabla_{\bm\theta} \log \oywbyisenw{\pi_{\bm\theta}}({a}_{t} | \bm{s}_{t})\right) \left(\sum_{t=1}^{N+1} r_t\right)\right]
\label{1616}
\end{align}
where $\mathbb E_\tau$ stands for $\mathbb E_{\oywbyisenw{\tau \sim \bm p_{\bm\theta}(\tau)}}$.
For a large enough batch $B$ of trajectories, \eqref{1616} \pw{is approximated with the empirical mean}: 
\begin{equation}
J(\bm\theta)\approx \hat{\mathbb E}_B\left[\sum_{t=1}^N r^{(b)}_{t}\right] = \frac{1}{|B|} \sum_{b=1}^{|B|} \sum_{t=1}^{N+1} r^{(b)}_{t} \label{1717}
\end{equation}
where $r^{(b)}_{t} = r(\bm{s}^{(b)}_{t}, {a}^{(b)}_{t})$,
$\bm{s}^{(b)}_{t}$ (resp. ${a}^{(b)}_{t}$) is the state (resp. action) at time step $t$ of the $b$-th trajectory $\tau^{(b)}$ generated by $\pi_{\bm\theta}$, and
$\hat{\mathbb E}_B$ denotes the empirical mean operation.
\pw{Then the policy gradient in \eqref{1616} can be approximated by:}
\begin{align}
\nabla_{\bm\theta} J(\bm\theta) \!\approx\! \hat{\mathbb E}_B\left[\big(\sum_{t=1}^{N} \nabla_{\bm\theta} \log \oywbyisenw{\pi_{\bm\theta}}({a}^{(b)}_{t} | \bm{s}^{(b)}_{t})\big)
\big(\sum_{t=1}^{N+1} r^{(b)}_{t}\big)\right]
\label{eq:pg}
\end{align}

\pw{Instead of updating $\bm\theta$ with this policy gradient, in our interleaved training, we use:
\begin{align}
- \hat{\mathbb E}_B\left[\big(\sum_{t=1}^{N} \nabla_{\bm\theta} \log \oywbyisenw{\pi_{\bm\theta}}({a}^{(b)}_{t} | \bm{s}^{(b)}_{t})\big)
L_{\sigma_+^{(b)}}(X)\right]
\label{eq:pgm}
\end{align}
where $\sigma_+^{(b)}$ is the improved tour obtained from our combined local search from $\sigma^{(b)}$, the tour induced by trajectory $\tau^{(b)} \in B$. 
By construction, $L_{\sigma_+^{(b)}}(X) \le L_{\sigma^{(b)}}(X) = -\sum_{t=1}^{N+1} r^{(b)}_{t}$.
}

\subsubsection{\Baseline{} baseline}\label{subsubsec:baseline}
\pw{In order to reduce the variance of the policy gradient estimate \eqref{eq:pgm}, we use a simple baseline and update $\bm\theta$ in the following direction:
\begin{align}
-\hat{\mathbb E}_{B}\left[\big(\sum_{i=1}^{N} \nabla_{\bm\theta} \log \oywbyisenw{\pi_{\bm\theta}}({a}^{(b)}_{t}|\bm{s}^{(b)}_{t})\big) 
\big( L_{\sigma_+^{(b)}}(X) - l^{(b)}  \big)\right]
\label{eq:RL_with_baseline}
\end{align}
where $l^{(b)} =-L_{\sigma^{(b)}}(X)$ is the baseline, which we call the \textit{\baseline} baseline.
Such a baseline gives more weight in the policy gradient when local search can make more improvement. 
In our experiments, our baseline performs better than the previous greedy baselines \cite{attention,GPN} in our training process. One other nice feature of our baseline is that it does not incur any extra computation since $L_{\sigma^{(b)}}(X)$ is already computed when policy generates $\sigma^{(b)}$.
}

\subsection{Stochastic Curriculum Learning}

\pw{Curriculum Learning (CL) is a widely-used technique in machine learning (and RL) \cite{CL_survey}, which can facilitate learning and improve generalization.
Although it can be implemented in various ways, its basic principle is to control the increasing difficulty of the training instances.}

\pw{To train \MAGIC{}, we} propose a {\bf stochastic CL} \pw{technique} where the probability of choosing harder instances increases over training steps.
\pw{We choose the number of cities as a measure of difficulty for a TSP instance, which is assumed to be in \pw{$\{10, 11, \ldots, 50\}$ in our experiments}.
We explain next how this selection probability is defined.
}

\pw{For} epoch $e$, we define the vector $\bm g^{(e)}\in\mathbb{R}^{41}$ (
since there are 41 integers between 10 and 50) to be 
\begin{equation}
 \bm g^{(e)}_{k}=\frac{1}{\sqrt{2 \pi }\pw{\sigma_{\mathcal N}}} \exp^{-\frac{1}{2}\left(\frac{(k+10)-e}{\sigma_{\mathcal N}}\right)^{2}}
\end{equation}
where $\bm g^{(e)}_{k}$ represents the $k^{th}$ entry of $\bm g^{(e)}$,
and $\sigma_{\footnotesize{\mathcal N}}$ is a hyperparameter which represents the standard deviation of the normal distribution.
Then, we use a softmax to formulate the probability distribution $\bm p^{(e)}$ of this epoch 
\begin{equation}
     \bm p^{(e)}=\text{softmax}( \bm g^{(e)})
\end{equation}
where $\bm p^{(e)}\in[0,1]^{41}$, and the $k^{th}$ entry of $\bm p^{(e)}$ 
represents to probability of choosing TSP of $(k+10)$ \pw{cities} at epoch $e$.

\subsection{\pw{Overall}
training process}
In this part, we \pw{summarize} our training process by providing the corresponding pseudo code in Algorithm \ref{alg:Training}.

\begin{algorithm}[tb]
  \caption{REINFORCE with Stochastic CL,  Policy Rollout Baseline and Combined local search}
  \begin{algorithmic}[1]
  \STATE {\bf Input}: Total number of epochs $E$, training steps per epoch $T$, batch size $|B|$, hyperparameters $\alpha$, $\beta$, $\gamma$ and $I$ for local search
  \STATE Initialize $\bm\theta$
  \FOR{$e=1$ \TO $E$}
    \STATE $N\leftarrow$ Sample from $\bm p^{(e)}$ according to Stochastic CL
    \FOR{$t=1$ \TO $T$}
    \STATE $\forall{b\in\{1,...,|B|\}}\,X^{(b)}\,\leftarrow$ Random TSP instance with $N$ cities
    \STATE $\forall{b\in\{1,...,|B|\}}\,\sigma^{(b)}\,\leftarrow$ Apply $\pi_{\bm\theta}$ on $X^{(b)}$
    \vspace{0.1cm}
    \label{alg:model}
    \STATE $\forall{b\in\{1,...,|B|\}}\,\sigma_+^{(b)}\,\leftarrow$ Apply the combined local search on $\sigma^{(b)}$
    \STATE Compute gradient in \eqref{eq:RL_with_baseline} using $\sigma^{(b)}$ and $\sigma_+^{(b)}$
    \STATE $\bm\theta\leftarrow$ Update in the direction of this gradient 
    \ENDFOR
  \ENDFOR
  \end{algorithmic}
\label{alg:Training}
\end{algorithm}
Notice that line~\pw{\ref{alg:model}} can be replaced by any model that can used to generate a \pw{tour}, showing that a variety of models can fit in our training process to improve their performance for TSP problems.

\section{Experiment\pw{s}} \label{sec:expe}
\begin{table*}[tb]
\caption{Results and comparisons on small TSP cases, obtained by testing on 10,000 instances for TSP 20, 50 and 100}
\label{Table:small}
\begin{tabular}{@{}lllllllllll@{}}
\toprule[1.2pt]
Method                   & Type                             & \multicolumn{3}{c}{TSP20} & \multicolumn{3}{c}{TSP50} & \multicolumn{3}{c}{TSP100} \\
\cmidrule(lr){3-5}
\cmidrule(lr){6-8}
\cmidrule(lr){9-11}
                         &                                  & Length & Gap     & Time(s) & Length & Gap     & Time(s) & Length  & Gap     & Time(s) \\ \midrule
Concorde$^{*}$                 & Exact Solver                     & 3.830  & 0.00\%  & 138.6   & 5.691  & 0.00\%  & 820.8   & 7.761   & 0.00\%  & 3744    \\
Gurobi$^{*}$                   & Exact Solver                     & 3.830  & 0.00\%  & 139.8   & 5.691  & 0.00\%  & 1572    & 7.761   & 0.00\%  & 12852   \\ \midrule
2-opt                     & Heuristic                        & 4.082  & 6.56\%  & 0.33    & 6.444  & 13.24\% & 2.25    & 9.100   & 17.26\% & 9.32    \\
Random Insertion         & Heuristic                        & 4.005  & 4.57\%  & 196     & 6.128  & 7.69\%  & 502.2   & 8.511   & 9.66\%  & 1039    \\
Nearest Insertion        & Heuristic                        & 4.332  & 13.10\% & 229.8   & 6.780  & 19.14\% & 633     & 9.462   & 21.92\% & 1289    \\
Farthest Insertion        & Heuristic                        & 3.932  & 2.64\%  & 239.8   & 6.010  & 5.62\%  & 617     & 8.360   & 7.71\%  & 1261    \\ \midrule
GCN$^{*}$ (Joshi et al.)        & SL (Greedy)                       & 3.855  & 0.65\%  & 19.4    & 5.893  & 3.56\%  & 120     & 8.413   & 8.40\%  & 664.8   \\
Att-GCRN+MCTS$^{*}$ (Fu et al.) & SL+\pw{MCTS}                            & 3.830  & 0.00\%  & 98.3    & 5.691  & 0.01\%  & 475.2   & 7.764   & 0.04\%  & 873.6   \\
GAT$^{*}$ (Deudon et al.)       & RL (Sampling)                     & 3.874  & 1.14\%  & 618     & 6.109  & 7.34\%  & 1171    & 8.837   & 13.87\% & 2867    \\
GAT$^{*}$ (Kool et al.)         & RL (Greedy)                       & 3.841  & 0.29\%  & 6.03    & 5.785  & 1.66\%  & 34.9    & 8.101   & 4.38\%  & 109.8   \\
GAT$^{*}$ (Kool et al.)         & RL (Sampling)                     & 3.832  & 0.05\%  & 988.2   & 5.719  & 0.49\%  & 1371    & 7.974   & 2.74\%  & 4428    \\
GPN (Ma et al.)           & RL                     & 4.074  & 6.35\%  & 0.77    & 6.059  & 6.47\%  & 2.50    & 8.885   & 14.49\% & 6.23    \\
\MAGIC{} (Ours)                                          & RL (Local Search)  & 3.870  & 1.09\%  & 3.06     & 5.918  & 4.00\%  & 14.8    & 8.256  & 6.39\%   & 50.4    \\ \bottomrule[1.2pt]
\end{tabular}
\begin{tablenotes}
    \footnotesize
    \item * refers to methods whose results we directly use from others' papers.
\end{tablenotes}
\end{table*}

\begin{table*}[tb]
\caption{Results and comparisons on large TSP cases, obtained by testing on 128 instances for TSP 200, 500 and 1000}
\label{Table:big}
\begin{tabular}{@{}lllllllllll@{}}
\toprule[1.2pt]
Method                   & Type                             & \multicolumn{3}{c}{TSP200}                       & \multicolumn{3}{c}{TSP500}                         & \multicolumn{3}{c}{TSP1000}                        \\
\cmidrule(lr){3-5}
\cmidrule(lr){6-8}
\cmidrule(lr){9-11}
                         &                                  & Length          & Gap             & Time(s)       & Length          & Gap              & Time(s)        & Length          & Gap              & Time(s)        \\ \midrule
Concorde$^{*}$                 & Solver                           & 10.719          & 0.00\%          & 206.4         & 16.546          & 0.00\%           & 2260           & 23.118          & 0.00\%           & 23940          \\
Gurobi$^{*}$                   & Solver                           & -               & -               & -             & -               & -                & -              & -               & -                & -              \\ \midrule
2-opt                     & Heuristic                        & 12.841          & 19.80\%         & 34.0          & 20.436          & 23.51\%          & 201.7          & 28.950          & 25.23\%          & 826.2          \\
Random Insertion         & Heuristic                        & 11.842          & 10.47\%         & 27.1          & 18.588          & 12.34\%          & 68.3           & 26.118          & 12.98\%          & 137.0          \\
Nearest Insertion        & Heuristic                        & 13.188          & 23.03\%         & 28.8          & 20.614          & 24.59\%          & 79.8           & 28.971          & 25.32\%          & 176.6          \\
Farthest Insertion        & Heuristic                        & 11.644          & 8.63\%          & 33.0          & 18.306          & 10.64\%          & 84.0           & 25.743          & 11.35\%          & 175.5          \\ \midrule
GCN$^{*}$ (Joshi et al.)       & SL (Greedy)                       & 17.014          & 58.73\%         & 59.1          & 29.717          & 79.61\%          & 400.2          & 48.615          & 110.29\%         & 1711           \\
Att-GCRN+MCTS$^{*}$ (Fu et al.) & SL+\pw{MCTS}                            & 10.814          & 0.88\%          & 149.6         & 16.966          & 2.54\%           & 354.6          & 23.863          & 3.22\%           & 748.3          \\
GAT$^{*}$ (Deudon et al.)       & RL (Sampling)                     & 13.175          & 22.91\%         & 290.4         & 28.629          & 73.03\%          & 1211           & 50.302          & 117.59\%         & 2262           \\
GAT$^{*}$ (Kool et al.)         & RL (Greedy)                       & 11.610          & 8.31\%          & 5.03          & 20.019          & 20.99\%          & 90.6           & 31.153          & 34.75\%          & 190.8          \\
GAT$^{*}$ (Kool et al.)         & RL (Sampling)                     & 11.450          & 6.82\%          & 269.4         & 22.641          & 36.84\%          & 938.4          & 42.804          & 85.15\%          & 3838           \\
GPN$^{*}$ (Ma et al.)      & RL              & -      & -       & -        & 19.605 & 18.49\% & -       & 28.471 & 23.15\%  & -       \\
GPN+2opt$^{*}$ (Ma et al.)  & RL+2opt         & -      & -       & -        & 18.358 & 10.95\% & -        & 26.129 & 13.02\%  & -       \\
GPN (Ma et al.)                       & RL                               & 13.278          & 23.87\%         & 2.5           & 23.639          & 42.87\%          & 7.13           & 37.849          & 63.72\%          & 18.35          \\
\MAGIC{} (Ours)                                          & RL (Local Search) &  11.539 & 7.65\%  & 69.9     & 18.098 & 9.38\%  & 207.8    & 25.542 & 10.49\%  & 487.8   \\ \bottomrule[1.2pt]
\end{tabular}
\begin{tablenotes}
    \footnotesize
    \item * refers to methods whose results we directly use from others' papers.
\end{tablenotes}
\vspace{-0.2cm}
\end{table*}

\label{set:exp1}
\begin{table*}[tb]
\caption{Ablation study on RL, CL, the \pw{\baseline} baseline and the combined local search. Testing on 10,000 instances \pw{for} TSP 20, 50 and 100, and 128 instances \pw{for} TSP 200, 500 and 1000.}
\label{Table:ablation}
\centering
\begin{tabular}{@{}lllllllllll@{}}
\toprule[1.2pt]
Method & \multicolumn{2}{c}{\textbf{Full version}} & \multicolumn{2}{c}{w/o RL} & \multicolumn{2}{c}{w/o CL} & \multicolumn{2}{c}{w/o baseline} & \multicolumn{2}{c}{w/o local search} \\
\cmidrule(lr){2-3} \cmidrule(lr){4-5} \cmidrule(lr){6-7} \cmidrule(lr){8-9} \cmidrule(lr){10-11}
                            & \multicolumn{1}{c}{Length}              & \multicolumn{1}{c}{Gap}                  & \multicolumn{1}{c}{Length}              & \multicolumn{1}{c}{Gap}             & \multicolumn{1}{c}{Length}              & \multicolumn{1}{c}{Gap}             & \multicolumn{1}{c}{Length}              & \multicolumn{1}{c}{Gap}                  & \multicolumn{1}{c}{Length}              & \multicolumn{1}{c}{Gap}                       \\ \midrule
TSP20                       & \textbf{3.871}      & \textbf{1.07\%}      & 3.9556          & 3.27\%          & 3.917           & 2.27\%          & 3.911                & 2.21\%               & 3.988                     & 4.10\%                    \\
TSP50                       & \textbf{5.957}      & \textbf{4.69\%}      & 6.1391          & 7.88\%          & 5.959           & 4.72\%          & 5.983                & 5.15\%               & 5.962                     & 4.76\%                   \\
TSP100                      & \textbf{8.302}      & \textbf{6.97\%}      & 8.5419          & 10.06\%         & 8.343           & 7.51\%          & 8.395                & 8.17\%               & 8.331                     & 7.35\%                   \\
TSP200                      & \textbf{11.567}     & \textbf{7.91\%}      & 11.9299         & 11.30\%         & 11.682          & 8.99\%          & 11.842               & 10.48\%              & 11.631                    & 8.50\%                   \\
TSP500                      & \textbf{18.321}     & \textbf{10.73\%}     & 18.9036         & 14.25\%         & 18.332          & 10.80\%         & 18.516               & 11.91\%              & 18.526                    & 11.97\%                   \\
TSP1000                     & \textbf{25.854}     & \textbf{11.84\%}     & 26.9936         & 16.76\%         & 25.954          & 12.27\%         & 26.188               & 13.28\%              & 26.505                    & 14.65\%                   \\ \bottomrule[1.2pt]
\end{tabular}
\end{table*}

To demonstrate the performance and  generalization ability of our proposed training process, \pw{we evaluate our method on randomly generated TSP instances and compare with}
other existing algorithms, \oywbyisenw{which cover different kinds of methods for completeness and include two exact solvers, four traditional heuristics, and seven learning based ones.
If the hardware and experiment setting of other papers are the same as ours, we will directly use their reported results on time and performance.
To ensure a fair comparison \pw{of} the \pw{runtimes} and \pw{performances}, all algorithms are executed on a computer with an Intel(R) Xeon(R) CPU E5-2678 v3 and a single GPU 1080Ti, and parallel computation \pw{is} utilized as much as possible for all the algorithms.}
Moreover, to show the power of RL, CL, our policy rollout baseline and the combined local search, we \pw{also} carry out \pw{an} ablation study of those four components.

\subsection{Data sets and hyperparameters}

\pw{We denote TSP $n$ the set of random TSP instances with $n$ cities.}
For the sake of consistency with other research work, \pw{the} coordinates \pw{of} cities \pw{are} randomly (from a uniform distribution) generated from $[0,1]^2$. 
\pw{
For training, the \pw{TSP} size $n$ varies from 10 to 50 decided by CL in every epoch. 
After training the model \pw{according to} our training process, we test our model on both small and large TSP instances. 
For the testing of small TSP problems, we test on 10,000 instances respectively for TSP 20, 50 and 100. 
For the testing of large TSP problems, we test on 128 instances respectively for TSP 200, 500 and 1000 to test the generalization ability.} 

For the hyperparameters of the training, we train for 200 epochs, and  we process 1000 batches of 128 instances every epoch. 
For the learning rate, we set it initially to be 0.001 with a 0.96 learning rate decay. 
For the hyperparameters of local search, we set $\alpha=0.5$, $\beta=1.5$, $\gamma=0.25$ and $I = 25$ after a quick hyper-parameter search. 
Those settings aim to train a model with fast speed and generalization ability. 
\oywbyisenw{For the model architecture, the aggregation function used in GNN is represented by a neutral network followed by a ReLU function on each entry of the output. 
Our MLP has an input layer with dimension 2, two hidden layers with dimension $H$ and $2H$ respectively, and an output layer with dimension $H$. 
Layers are fully connected and we choose to use the ReLU as the activation function.
And finally, we set $H = 128$, $n_{GNN} = 3$.}

\subsection{Performance \pw{on} small-scale and large-scale TSP \pw{instances}}

The results are shown in \pw{Tables~\ref{Table:small} and} \ref{Table:big}. 
\pw{Column 1 and } 2 \pw{respectively specify the method and its} type, where SL refers to supervised learning, Greedy means a greedy construction from the probability given by the policy and Sampling refers to sampling multiple solutions from the probability given by the policy and choose the best one. 
\pw{Column} 3 indicates the average tour length, 
\pw{Column 4 provides the gap to Concorde's performance, which corresponds to the \oywbyisenw{optimal solution for the small-scale TSP problems and nearly optimal solution for the large-scale TSP problems}}, and 
\pw{Column} 5 \pw{lists} the total runtime. 
For comparisons, we have listed out 12 other methods covering from exact solvers, heuristics to learning-based algorithms. 

As shown in Table \ref{Table:small}, the exact solvers have the best performance but with relatively long runtime; most of the learning based methods, including ours, receive better tour length than heuristics. Within the learning based methods, most methods are not more than $2\%$ better than ours. For TSP 100, only \pw{\textcite{Fu}} and those who applied a sampling methods, which all use methods to search for best tour, have better tour than ours. For the speed, our method is fast and this is more prominent when the size is bigger. Those learning-based methods with better solutions than ours all run slower and only \pw{\textcite{GPN}} is faster but with a significantly worse solution than ours.   
For the results in Table \ref{Table:big}, the solver also gives the best solution while its speed is relatively slow; many learning-based methods now expose their poor generalization ability and give worse results than heuristics. 
For heuristics, insertions methods show a good performance on large TSP problems. 
For our methods, we outperform all the learning-based models except for \pw{\textcite{Fu}} for TSP 500 and 1000, showing a very good generalization ability. 
Plus, for the runtime, we are generally fast and especially faster than \pw{\textcite{Fu}}.

Notice that heuristics has a good generalization ability and previous learning-based algorithms do well in small TSP problems. Our learning based method combined with local search, which is inspired by the heuristics, tends to receive the advantage of learning based methods and heuristics. Plus, \pw{it is} also fast \pw{in terms of}  runtime, making it a comprehensive excellent method.

\subsection{The ablation study}
\label{set:exp2}

To demonstrate the importance of RL, CL, the \pw{\baseline} baseline, and the combined local search in the training process, we perform \pw{an} ablation study on them.
For \pw{this} study, since we only need to show the importance of \pw{each technique} we apply, we turn down the hyper-parameter $I = 15$ to have a shorter runtime.
For the ablation study on RL, we \pw{do not perform} any learning and directly apply the combined local search on \pw{randomly} generated tours. 
For \pw{the} ablation study on CL, we follow the same settings except that the \pw{TSP} size is fixed to be 50 for all epochs. 
For the ablation study on \pw{the \baseline} baseline, we use \pw{instead} the central self-critic baseline \cite{GPN}, which is inspired by the self-critic training  \cite{self-critics} and the greedy self-critic baseline \cite{attention}. 
For \pw{the} ablation study on the combined local search, we \pw{do not perform} any local search \pw{during} training, but we still apply it in testing in order to show \pw{that interleaving} local search \pw{with policy gradient updates} outperforms only doing post-optimization.
\pw{Note} that the \pw{\baseline} baseline depends on the combined local search. Therefore, since the local search is ablated for this study, the \pw{\baseline} baseline also needs to be changed, and here we replace it by the central self-critic baseline \cite{GPN}. 
\pw{The results of the ablation study presented in Table~\ref{Table:ablation} demonstrate that all the components used in our method contribute to its good performance.}




\section{Conclusion} \label{sec:conclusion}

We introduced a novel deep RL approach for solving TSP achieving state-of-the-art results in terms of gap to optimality and runtime compared to previous deep RL methods.
The results are particularly promising in terms of generalizability.
Our proposition consists of a simple deep learning architecture combining graph neural network, multi-layer perceptron, and attention in addition to a novel RL training procedure, which interleaves local search and policy gradient updates, uses a novel return evaluation, and exploits curriculum learning.

As future work, we plan to evaluate our novel training procedure with other deep learning architectures proposed for TSP, but also other combinatorial optimization problems.
Another research direction to improve further our results is to optimize the deep learning architecture in order to improve the encoding of the TSP problem.

\section*{ACKNOWLEDGMENT}
This work is supported in part by the Innovative Practice Program of Shanghai Jiao Tong University (IPP21141), the program of National Natural Science Foundation of China (No. 61872238), and the program of the Shanghai NSF (No. 19ZR1426700).
Moreover, we thank \textcite{GPN} for sharing their source code, which served as initial basis for our work.

\printbibliography

\end{document}